\def\year{2021}\relax
\documentclass[letterpaper]{article} 
\usepackage{aaai21}  
\usepackage{times}  
\usepackage{helvet} 
\usepackage{courier}  
\usepackage[hyphens]{url}  
\usepackage{graphicx} 
\usepackage{natbib}  
\usepackage{caption} 
\usepackage{amsmath}
\usepackage{booktabs,makecell,multirow,array,caption}
\usepackage{csquotes}
\usepackage{cleveref}
\usepackage[dvipsnames]{xcolor}

\colorlet{LightSpringGreen}{White!50!SpringGreen}
\colorlet{LightSkyBlue}{White!50!SkyBlue}
\colorlet{LightBrickRed}{White!50!Salmon}
\title{Efficient Retrieval Augmented Generation\\from Unstructured Knowledge for Task-Oriented Dialog}
\author{
    David Thulke,\textsuperscript{\rm 1,2}
    Nico Daheim,\textsuperscript{\rm 1}
    Christian Dugast,\textsuperscript{\rm 1,2}
    Hermann Ney\textsuperscript{\rm 1,2}\\
}
\affiliations{
    \textsuperscript{\rm 1} Human Language Technology and Pattern Recognition Group, RWTH Aachen University, Germany \\
    \textsuperscript{\rm 2} AppTek GmbH, Aachen, Germany \\
    \{thulke, daheim, dugast, ney\}@i6.informatik.rwth-aachen.de
}

\begin{document}

\maketitle

\begin{abstract}
    This paper summarizes our work on the first track of the ninth Dialog System Technology Challenge (DSTC 9), \enquote{Beyond Domain APIs: Task-oriented Conversational Modeling with Unstructured Knowledge Access}.
    The goal of the task is to generate responses to user turns in a task-oriented dialog that require knowledge from unstructured documents.
    The task is divided into three subtasks: detection, selection and generation.
    In order to be compute efficient, we formulate the selection problem in terms of hierarchical classification steps. 
    We achieve our best results with this model.
    Alternatively, we employ siamese sequence embedding models, referred to as Dense Knowledge Retrieval, to retrieve relevant documents.
    This method further reduces the computation time by a factor of more than 100x at the cost of degradation in R@1 of 5-6\% compared to the first model.
    Then for either approach, we use Retrieval Augmented Generation to generate responses based on multiple selected snippets and we show how the method can be used to fine-tune trained embeddings.
\end{abstract}

\section{Introduction}

Task-oriented dialog systems allow users to achieve certain goals, such as restaurant or hotel reservations, by interacting with a system using natural language.
Typically, these systems are restricted to information provided by an application-specific interface (API) which accesses a structured database.
However, the breadth of structured information available is often limited to a set of fixed fields and may not satisfy all information needs users may have.
The necessary information is in many cases already available but only found in unstructured documents, such as descriptions on websites, FAQ documents, or customer reviews.
The aim of Track 1 of the ninth Dialog System Technology Challenge (DSTC 9) \cite{DSTC9} is to make use of such information to provide users with an answer relevant to their question. 
To do so, the task at hand is split up into three subtasks, namely Knowledge-seeking \emph{Turn Detection} to identify those questions that can not be answered by an existing API, \emph{Knowledge Selection} to retrieve relevant documents, and \emph{Response Generation} to generate a suitable system response.

In a real-world scenario, the amount of time that can be spent on generating a response is limited.
Since the number of relevant documents can potentially grow large, we note that efficient means of retrieval are crucial.
In the baseline approach, this is identified as the limiting factor.
Thus, our work focuses on the task of Knowledge Selection and, specifically, efficient document retrieval methods.
For this, we propose two different methods.
Our first method splits the task of classifying a relevant document into subsequent stages.
Thereby we are not only able to significantly reduce the computational cost, but also to outperform the single-stage classification approach by a significant margin on both test sets.
Secondly, we employ siamese embedding networks to learn dense document and dialog embeddings, which are then used to retrieve relevant documents based on vector similarity. 
Additionally, we fine-tune these models jointly with a retrieval augmented generation model, achieving results comparable to the baseline.
This model can then also be used during generation, to condition the model on multiple documents.

\section{Related Work}

Recently, the use of pre-trained transformer language models like GPT-2 has shown to be successful for task-oriented dialog systems.
It is used either as a model in one of the subtasks, like response generation, or as an end-to-end model for the whole task \cite{budzianowskiHelloItGPT22019,hamEndtoEndNeuralPipeline2020}.
Other transformer models which have been pre-trained on other tasks such as BERT or RoBERTa have been shown to outperform GPT-2 on a range of natural language understanding tasks \cite{devlinBERTPretrainingDeep2019,liuRoBERTaRobustlyOptimized2019}.
Recently, \citet{lewisBARTDenoisingSequencetoSequence2020} proposed BART, an encoder-decoder model pre-trained as a denoising autoencoder.
On natural language understanding tasks like GLUE and SQuAD, it achieves similar performance to RoBERTa and can be effectively fine-tuned to sequence generation tasks, for example summarization or dialog response generation.

Most previous works considered the problem of integrating unstructured knowledge into conversations on open domain dialogs.
Existing benchmarks for example include conversations on topics grounded on Wikipedia articles \cite{dinanWizardWikipediaKnowledgePowered2018} or chats on movies \cite{mogheExploitingBackgroundKnowledge2018}.
\citet{ghazvininejadKnowledgeGroundedNeuralConversation2018} propose an encoder-decoder model that uses separate encoders, one for the dialogue context and one for knowledge documents.
Relevant documents are retrieved by named entity and keyword matching.
\citet{kimSequentialLatentKnowledge2020} jointly model knowledge selection and response generation using a sequential knowledge transformer.
\citet{zhaoKnowledgeGroundedDialogueGeneration2020} suggest an unsupervised approach for retrieval which does not require annotations of relevant knowledge documents.
Unlike in the current track of DSTC~9, in these benchmarks, typically, multiple knowledge documents are relevant to create a response.
Additionally, conversations in open domain dialogs often lack a clear goal and the state of the dialog is less constrained by the domain.

\section{Task Description}

In a task-oriented dialog system, the task is to generate an appropriate system response \(u_{T+1}\) given a user utterance \(u_T\) and a preceding dialog context \(u_1^{T-1}\).
In the context of the challenge \cite{kimDomainAPIsTaskoriented2020}, it is assumed that every user utterance can either be handled by an already existing API-based system or that it can be answered based on a set of knowledge documents \(K = \{k_1, \dots, k_N\}\).
These knowledge documents have different types of metadata, which shall be briefly introduced.
First of all, each document is uniquely assigned to one domain, e.g. restaurant or taxi, and one entity, e.g. a specific restaurant or hotel.
The set of all domains is denoted by \(D\) in the following. 
Each domain contains a set of entities \(E_d \subset E\), the set of all entities, for \(d \in D\).
Finally, each entity $e \in E_d$ is described by a set of knowledge documents \(K_e \subset K\).
We further define a special entity \(\ast_d\) for each domain \(d\) so that \(K_{\ast_d}\) contains all documents which are relevant to the whole domain.
The task of generating an appropriate system response based on knowledge is split into three subsequent subtasks, which we formally introduce in the following paragraphs.

\paragraph{Turn Detection}
In turn detection, the task of the system is to decide whether a turn should be handled by an existing API or by accessing an unstructured knowledge document.
Thus, the problem is formulated as a binary classification task, such that
\[
	f_1(u_1^T \mid K) = \begin{cases}
					1 & \text{if }\exists k \in K \text{ s.t. } k \text{ answers }u_T \\
					0 & \text{otherwise}
				   \end{cases}
\]
Furthermore, as noted in \citet{kimDomainAPIsTaskoriented2020}, each turn is assumed to be manageable by either API or unstructured knowledge, i.e. exactly one applies in each case.

\paragraph{Knowledge Selection}
After the turn detection, utterances that are classified as requiring unstructured knowledge access are handled in the knowledge selection task.
Here, the model has to retrieve one or more knowledge snippets relevant to the query posed by the last user turn and dialog context.
Hence, the task can be defined over the full dialog context and the full set of knowledge snippets as
\[
	f_2(u_1^T \mid K) = \{k \mid k \in K \land k \text{ relevant to } u_1^T\}
\]
In the following, we refer to the set of selected knowledge as \(\tilde{K} \subset K\).

\paragraph{Response Generation}
As a final step, the user is provided with a generated system's response based on the dialog context \(u_1^T\) and the set of selected knowledge snippets relevant to it, denoted by \(\tilde{K}\).
Thus, the task can be defined as
\[
	f_3(u_1^T, \tilde{K}) = u_{T+1} 
\]
Aside from containing the relevant knowledge, the generated response should naturally fit in the preceding dialog context \(u_1^T\).

\section{Methods}

In this section, we first briefly introduce the baseline methods suggested by \citet{kimDomainAPIsTaskoriented2020} and our modifications to them and then discuss additional extensions and alternatives proposed by us.
In all methods, we use pre-trained transformer models like RoBERTa or BART.
For classification tasks, we add a single feed-forward layer on top of the final layer output of the token representing the whole sequence (i.e. [CLS] for BERT or \texttt{\textless s\textgreater} for RoBERTa).
Due to the length limitations of these models and due to memory limitation of our hardware the input may be truncated to a maximum length, for example, by truncating the first utterances of the complete dialogue to reach the maximum input length.

\subsection{Baselines}
\label{sec:method_baseline}

\paragraph{Turn Detection}
To solve this subtask, \citeauthor{kimDomainAPIsTaskoriented2020} propose to train a binary classifier on the full dialog context \(u_1^T\).
We add a sigmoid to the single class output of the feed-forward layer and fine-tune the whole model using binary cross-entropy.

\paragraph{Knowledge Selection}
Similarly, \citeauthor{kimDomainAPIsTaskoriented2020} propose to model knowledge selection as a binary classification problem.
For a dialog context \(u_1^T\) and a knowledge snippet \(k\) the model predicts whether the knowledge snippet is relevant to the dialog or not.
It uses the same model architecture and training criterion as the previous task.
Instead of including all knowledge snippets that are not relevant to a dialog context into the loss, a random subset is sampled to make the training efficient.
During evaluation, this approximation is not possible and we have to consider all \(|K|\) knowledge snippets.
The method returns the knowledge snippet with the highest relevance score as the output of this task.

Unlike the baseline implementation, we augment the representation of the knowledge snippet with the name of the associated domain which is automatically taken from the snippet meta-data.
This especially reduces the confusion between domains that contain similar documents.
See \Cref{sec:evaluation} for more details.

\paragraph{Response Generation}
Finally, for response generation, \citeauthor{kimDomainAPIsTaskoriented2020} suggest to use an autoregressive sequence generation model conditioned on the dialog context and the knowledge snippet selected in the previous subtask.
\citeauthor{kimDomainAPIsTaskoriented2020} propose to fine-tune a pre-trained GPT-2 model \cite{radford2019gpt2} for this task.
We fine-tune a pre-trained BART model \cite{lewisBARTDenoisingSequencetoSequence2020} which is, unlike GPT-2, an encoder-decoder model.
In contrast to \citeauthor{kimDomainAPIsTaskoriented2020} who use Nucleus Sampling \cite{holtzmanCuriousCaseNeural2020} for decoding, we use beam search with a beam size of 4.
Additionally, we set a repetition penalty of 1.2 \cite{keskarCTRLConditionalTransformer2019} and a maximum decode length of 60 tokens.

\subsection{Hierarchical Selection}

One of the main issues with the baseline approach for knowledge selection is its computational complexity.
In the following, we refer to \(|D|\), \(|E|\) and \(|K|\) as the total number of domains, entities and knowledge snippets.
To calculate a relevance score for each knowledge snippet, \(|K|\) inference computations using the whole model are required.
In a real-time dialog application, this may become prohibitive.

An approach to solve this issue is to make use of the additional metadata available for each knowledge snippet.
Instead of directly selecting the relevant knowledge snippet from all available documents, we can divide the problem by first identifying the relevant domain and entity and then selecting the relevant snippet among the documents of this entity.
Therefore we use the same relevance classification model as in the baseline.
We try two variants of this approach.
In the first one, we train three separate models to retrieve the relevant domain, entity, and document.
In the second one, we train one model to jointly retrieve the relevant domain and entity and one model to retrieve the relevant document.

In total the approach reduces the complexity of the task from
\[
    \mathcal{O}\left(|K|\right) = \mathcal{O}\left(|D| \cdot \frac{|E|}{|D|} \cdot \frac{|K|}{|E|}\right)
\]
to
\[
    \mathcal{O}\left(|D| + \frac{|E|}{|D|} + \frac{|K|}{|E|}\right) \text{ and } \mathcal{O}\left(|E| + \frac{|K|}{|E|}\right)
\]
for the first and second variants.

\subsection{Dense Knowledge Retrieval}

While this approach decreases computational complexity, it may still be infeasible if the number of knowledge snippets gets too large.
\Citeauthor{kimDomainAPIsTaskoriented2020} showed that classical information retrieval approaches like TF-IDF and BM25 are significantly outperformed by their relevance classification model.
Simple bag-of-words models do not seem to be able to capture the relevant information of a dialog context necessary to retrieve the correct document.

Recent approaches like Sentence-BERT \cite{reimersSentenceBERTSentenceEmbeddings2019} or Dense Passage Retrieval \cite{karpukhinDensePassageRetrieval2020} showed that sentence representations based on pre-trained transformer models can be used effectively in information retrieval tasks when they are fine tuned on a task.
Similar to them we propose to use a siamese network structure made of a dialog context encoder and a knowledge snippets encoder so that the distance between their representation builds a suitable ranking function.
Thus, this essentially becomes a metric learning problem.
For the encoders, we use pre-trained RoBERTa \cite{liuRoBERTaRobustlyOptimized2019} models.
We directly use the last hidden state of the \texttt{\textless s\textgreater} token.
We experiment with two different distance functions: the euclidian distance and the dot product.

For the former, we use the triplet loss \cite{weinberger2009distance}, to train both encoders.
Given an anchor, in our case the encoded dialog context, and a positive and negative sample, in our case the relevant and a random irrelevant knowledge snippet, it trains the encoders so that the distance between the anchor and positive sample is lower than the distance to a negative snippet by a margin \(\epsilon\).

For the second method, we use the dot product between the embeddings created by encoder \(E_1\) and \(E_2\) as a similarity measure.
We train the model, given an anchor \(a\), to correctly classify a positive sample given the positive sample \(p\) and a set of negative samples \(N\).
Mathematically the loss is the negative log-likelihood of the correct positive sample:
\[
    L = - \log \frac{\exp{\left(E_1(a) \cdot E_2(p)\right)}}{\sum_{s \in N \cup \{p\}} \exp{\left(E_1(a) \cdot E_2(s)\right)}}    
\]
The anchor can either be a dialog context and the other samples are relevant and irrelevant knowledge snippets or the other way around.
To select negative examples we experiment with two different batching strategies.
The first is to randomly sample among the full set of negative samples.

Alternatively, instead of randomly sampling negative examples, one approach in metric learning tasks is to use hard negatives, i.e. the samples from the negative class whose embeddings are closest to the anchor.
To implement this efficiently, either only samples in the current batch are considered or the hardest samples are selected from a random subset.
For simplicity, we use the second approach in this work.

Since the embeddings of the knowledge snippets can be pre-computed only the embedding of the current dialog context has to be computed during runtime.
If the total number is rather small, i.e. in the thousands as in our case, the k nearest neighbor search to find the closest embedding is negligible compared to the inference time of the transformer model.
Thus, effectively this method is in \(\mathcal{O}(1)\).
Even for a very large number of knowledge snippets there are efficient means of search \cite{johnsonBillionscaleSimilaritySearch2019}.

\subsection{Retrieval Augmented Generation}
\label{sec:rag}

The baseline approach for response generation only considers the single best selected knowledge snippet.
In some cases, multiple snippets might contain relevant information to generate a response.
Further, by making a hard decision for a single knowledge snippet in the selection step, we introduce errors that are propagated to the response generation.
This motivates us to reformulate our selection and generation task into a single task, i.e. to generate a response based on all of our knowledge snippets.
The approach is similar to what \citet{lewisRetrievalAugmentedGenerationKnowledgeIntensive2020} propose and to other retrieval augmented models like REALM \cite{guuREALMRetrievalAugmentedLanguage2020}.
Mathematically, we can formulate this as a marginalization over the selected knowledge snippet \(k\) which we introduce as a latent variable:
\[
    p(u_{T+1} | u_1^T; K) = \sum_{k \in K} p(u_{T+1}, k | u_1^T; K)
\]
which can then be further split into a selection probability, i.e. the probability of a knowledge snippet given a dialog context, and a generation probability which corresponds to the baseline model for generation:
\[
    p(u_{T+1}, k | u_1^T; K) = \underbrace{p\left (k \mid u_1^T ; K \right )}_\textrm{selection} \cdot \overbrace{p\left (u_{T+1} \mid u_1^T, k ; K\right )}^{\textrm{generation}}
\]
The same decomposition can also be applied on the token level which allows us to use this as a drop-in replacement for our current generation probability.
To be able to calculate this efficiently during training and testing, we approximate the sum over all knowledge snippets \(K\) by a sum over the top \(n\) snippets.
To ensure that the model is still normalized, we renormalize the selection probabilities over this subset.
In our experiments, we typically use \(n=5\) and ensure that the correct knowledge snippet is always included in the top \(n\) snippets during training.
For the generation probability, we use the same model as in the baseline.
For the selection probability, we try all three models discussed so far.
In theory, this model allows us to train the selection and generation models jointly.
However, calculating the selection probabilities using the relevance classification models during training on the fly is not feasible, even when using the Hierarchical Selection models.
Therefore we calculate these probabilities in a previous step and keep them fixed during training.

Fortunately, using the Dense Knowledge Retrieval model, training both models jointly becomes feasible.
Therefore, we keep the knowledge snippet encoder fixed and only fine-tune the dialog context encoder.
The top \(n\) knowledge snippets can then be effectively retrieved during training.

\subsection{Multi-Task Learning}

Motivated by the recent success of multi-task learning for various NLP tasks \cite{liuMultiTaskDeepNeural2019,raffelExploringLimitsTransfer2020}, we explored two approaches to apply the method in this challenge.
In the first approach, we apply it to the Hierarchical Selection method.
Instead of training three separate models (or two if we classify the domain and entity jointly), we train a single model on all three relevance classification tasks.
For the second approach, we train a single model on all three subtasks of the challenge.
In both scenarios, we employ hard parameter sharing where the hidden layers are shared among all tasks \cite{ruder2017overview}.
On top of the shared layers, we add separate feed-forward layers for each task on top of the final layer output of the [CLS] token, to obtain separate outputs for each task.
For each relevance classification task we only include the relevant parts of the knowledge snippet into the input, i.e. to predict the relevant entity we only include the domain and the name of the entity but not the content of the document and calculate a loss based on the relevant output.
We used the same strategies to sample positive and negative samples as for the baseline models.

\begin{table*}
    \caption{Main results of our approaches on the \emph{validation} data. No. corresponds to the entry ID in DSTC 9.}
    \label{table:main_results_val}
    \centering
    \begin{tabular}{r|l|r|l|r|l|r|r|r}
        \hline
        no. & \multicolumn{1}{l}{detection}  &                & \multicolumn{1}{l}{selection}       &               & \multicolumn{3}{l}{generation}  \\
            & model                          & F1             & model                               & R@1           & model                   & BLEU-1        & METEOR        & ROUGE-L \\ \hline \hline
        -   & baseline                       & 98.5           & baseline                            & 67.3          & baseline                & 36.0          & 36.0          & 35.0  \\ \hline \hline
        -   & baseline (our)                 & 99.1           & baseline (our)                      & 91.1          & baseline (our)          & 42.8          & 43.5          & 41.0 \\ \cline{1-5}\cline{7-9}
        2   & \multirow[t]{5}{*}{Multi-task} & \textbf{99.7}  & \multirow[t]{2}{*}{Multi-Task}      & 94.1          &                         & \textbf{43.2} & 44.0          & 41.4 \\ \cline{1-1}\cline{3-3}\cline{5-9}
        1   &                                & \textbf{99.7}  &                                     & 94.1          & \multirow[t]{4}{*}{RAG} & 43.0          & 44.1          & 41.5 \\ \cline{1-1}\cline{3-5}\cline{7-9}
        3   &                                & \textbf{99.7}  & Hierarchical                        & \textbf{96.1} &                         & \textbf{43.2} & \textbf{44.5} & \textbf{41.6}  \\\cline{1-1} \cline{3-5}\cline{7-9}
        0   &                                & \textbf{99.7}  & DKR NLL                             & 93.2          &                         & 42.7          & 44.1          & 41.4 \\ \cline{1-1}\cline{3-5}\cline{7-9}
        -   &                                & \textbf{99.7}  & DKR Triplet                         & 90.7          &                         & 39.6          & 41.3          & 38.6 \\ \hline
    \end{tabular}
\end{table*}

\begin{table*}
    \caption{Main results of our approaches on the \emph{test} data. No. corresponds to the entry ID in DSTC 9.}
    \label{table:main_results_test}
    \centering
    \begin{tabular}{r|l|r|l|r|l|r|r|r}
        \hline
        no. & \multicolumn{1}{l}{detection}  &                & \multicolumn{1}{l}{selection}       &               & \multicolumn{3}{l}{generation}  \\
            & model                          & F1             & model                               & R@1           & model                   & BLEU-1        & METEOR        & ROUGE-L \\ \hline \hline
        -   & baseline                       & 94.5           & baseline                            & 62.0          & baseline                & 30.3          & 29.8          & 30.3  \\ \hline \hline
        -   & baseline (our)                 & 96.1           & baseline (our)                      & 87.7          & baseline (our)          & \textbf{38.3}  & 38.5          & 37.1 \\ \cline{1-5}\cline{7-9}
        2   & \multirow[t]{5}{*}{Multi-task} & \textbf{96.4}  & \multirow[t]{2}{*}{Multi-Task}      & 78.7          &                         & 37.5      & 38.1          & 36.8 \\ \cline{1-1}\cline{3-3}\cline{5-9}
        1   &                                & \textbf{96.4}  &                                     & 78.7          & \multirow[t]{4}{*}{RAG} & 37.5          & 38.0         & 36.7 \\ \cline{1-1}\cline{3-5}\cline{7-9}
        3   &                                & \textbf{96.4}  & Hierarchical                        & \textbf{89.9} &                         & 37.9      & \textbf{38.6} & 37.1  \\\cline{1-1} \cline{3-5}\cline{7-9}
        0  &                                & \textbf{96.4}  & DKR NLL                             & 83.8          &                         & 38.1          & 38.4          & \textbf{37.3} \\ \cline{1-1}\cline{3-5}\cline{7-9}
        -   &                                & \textbf{96.4}  & DKR Triplet                         & 84.4          &                         & 35.2          & 37.0          & 35.6 \\ \hline
    \end{tabular}
\end{table*}

\section{Data}

The training data provided as part of the challenge for this task is based on the MultiWOZ 2.1 dataset \cite{budzianowskiMultiWOZLargeScaleMultiDomain2018,ericMultiWOZConsolidatedMultiDomain2020}.
MultiWOZ is a task-oriented dialog dataset consisting of 10,438 dialogs spanning 7 different domains.
Each of these domains is defined by an ontology and has a database of corresponding entities.
For this challenge, \citet{kimDomainAPIsTaskoriented2020} extended the corpus by adding user turns requesting information not covered by the database, corresponding system responses, and knowledge snippets for each entity.
The latter were collected from the FAQ websites of the corresponding entities occurring in the corpus. Each snippet consists of a domain, an entity, a question, and an answer.
The additional turns were collected with the help of crowd workers.
In total, 21.857 new pairs of turns and 2.900 knowledge snippets were added to the corpus. 
The training and validation datasets are restricted to four domains, namely hotel, restaurant, taxi, and train.
The latter two domains do not contain any entities and corresponding knowledge snippets are relevant for the whole domain.

The organizers of the challenge announced that the final test data will include additional new domains, entities, and knowledge snippets.
To simulate these conditions when evaluating our approaches, we created an additional training dataset in which we removed all dialogs corresponding to the train and restaurant domain.

The final test data introduced one additional domain: attraction.
It contains 4.181 additional dialogs of which 1.981 have knowledge-seeking turns and 12.039 knowledge snippets.
Around half of these dialogs are from the MultiWOZ dataset augmented as described above.
The other half are human-to-human conversations about touristic information for San Francisco.
Of these, 90\% are written conversations and 10\% transcriptions of spoken conversations.

\section{Experiments}
\label{sec:evaluation}

We implemented our proposed methods on top of the baseline implementation provided as part of the challenge\footnote{Code is available at \url{https://github.com/dthulke/dstc9-track1}}.
For the pre-trained models, we used the implementation and checkpoints provided by Huggingface's Transformer library \cite{Wolf2019HuggingFacesTS}.
We use RoBERTa \cite{liuRoBERTaRobustlyOptimized2019} as a pre-trained language model for the encoders of the embedding models.
In all other cases, if not otherwise stated, we use BART large \cite{lewisBARTDenoisingSequencetoSequence2020}.
For the selection, dialog contexts are truncated to maximum length of 384 tokens.

To organize our experiments we used the workflow manager Sisyphus \cite{peterSisyphusWorkflowManager2018}.
All models are trained on Nvidia GTX 1080 Ti or RTX 2080 Ti GPUs.

\subsection{Results}

\Cref{table:main_results_val,table:main_results_test} show our main results on the validation and test data for all three tasks.
The results of the selection and generation are based on the results of the previous subtasks.
In all other tables, these results are based on the ground truth of the previous task to increase comparability.
The first two lines in the table for each evaluation dataset compare the results of the baseline method of \citet{kimDomainAPIsTaskoriented2020} and using our general modifications discussed in \Cref{sec:method_baseline}.
Rows with entry IDs correspond to the systems we submitted to DSTC 9 as Team 18.
We achieved our best result with system 3 with which we ranked on the 6th place in the automatic and on the 7th place in the human evaluation.

The next section analyzes the effect of these different modifications in more detail.
In the remainder of this subsection, we discuss the results of our proposed methods on each subtask.

\subsubsection{Turn Detection}
In the detection task, we achieve our best results with a \emph{Multi-Task} model trained on the three losses of each task.
Compared to the baseline, this approach mainly improved the recall of the model on the validation and test data with a slight degradation of the precision.
One possible explanation for this improvement is that joint training helps the model to focus on relevant parts of the dialog context.
We decided to use the Multi-Task detection as our standard method for this task.

\begin{table}
    \caption{Runtimes in seconds per turn for different methods on one GTX 1080 Ti.}
    \label{table:latencies}
    \centering
	\begin{tabular}{l|l|r|r}
	\hline
	task       & model 				 & \multicolumn{2}{c}{runtime} \\
		   	   &				 	 & validation	   & test   \\ \hline
	detection  & baseline 		 & 0.04    & 0.04   \\ \hline
	selection  & baseline 		 & 111.66  & 276.53 \\
		   	   & Hierarchical		 & 4.60    & 13.79  \\
	   	       & DKR				 & 0.04    & 0.04   \\\hline
	generation & baseline			 & 0.85    & 0.82   \\
		       & RAG + DKR 			 & 1.20    & 1.48   \\ \hline	

	\end{tabular}
\end{table}

\subsubsection{Knowledge Selection}
The \emph{Hierarchical Selection} model achieves the best results for MRR@5 and R@1 of all selection models we tested.
Other models outperform the model concerning R@5.
This can be explained by the fact that the model only returns documents of a single entity in its final ranking, thus these numbers are not fully comparable.
When analyzing the improvements in detail, we mainly see that the number of confusions among similar documents of different entities reduces.
Due to the hierarchical structure, the model is forced to first decide which domain and entity is relevant.
The other models have to make a tradeoff with respect to the relevance of the document itself and the relevance of the domain and entity.
Furthermore, Hierarchical Selection generalizes very well to new domains and sources (R@1 of 98.5 for attraction, 94.4 for sf\_written, and 87.5 for sf\_spoken).
As expected, it achieves a significant speedup of 20x compared to the baseline selection method as shown in \Cref{table:latencies}.
Even so, for a real-time application, a latency of around 13 seconds is still too high.

The \emph{Dense Knowledge Retrieval (DKR)} model achieves an additional speedup of more than a factor of 100x compared to the hierarchical selection model and more than a factor of 2,500x compared to the baseline model.
On the validation data, we observed that the negative log-likelihood (NLL) loss outperforms the triplet loss and even achieves better results than the baseline method.
Nevertheless, the model trained using the triplet loss seems to generalize better to the test data where it outperforms the model trained using the NLL loss.
As shown in \Cref{table:selection}, the performance of the models is significantly improved by joint training with the RAG model.
One interesting observation is that the DKR models do not generalize well to the spoken data (R@1 goes down to 43.2 for the DKR NLL model) compared to other models.

Finally, we trained a \emph{Multi-Task} model on all tasks of the Hierarchical Selection model and used it to calculate relevance scores for all documents, similar to the baseline approach.
On the validation data this model outperformed the baseline and DKR models but gave worse results on the test data.

\subsubsection{Response Generation}
As one can see from the results of entry ids 1 and 2 in \Cref{table:main_results_val,table:main_results_test}, there are no significant improvements by using Retrieval Augmented Generation over the baseline method in terms of automatic evaluation metrics.
Nevertheless, \Cref{table:rag} illustrates two examples, where it is beneficial.
As mentioned in \Cref{sec:rag}, RAG can incorporate multiple knowledge snippets into the generated response.
In the first row, for example, the baseline method fails to mention dry cleaning, even though it was specifically mentioned in the last user turn.
RAG, however, can give information on both laundry and dry cleaning.
Furthermore, by conditioning the probability of a generated response on the selected snippet, errors made in the selection can be corrected using RAG and are not necessarily propagated, as shown in the second example of the table.
Thus, all of our systems shown in this section, except for the baseline models, use Retrieval Augmented Generation.

\subsection{Ablation Analysis}

We perform an ablation study to analyze the influence of different changes on the results summarized above.
In contrast to the results in \Cref{table:main_results_val,table:main_results_test}, all results discussed in this section are based on the ground truth labels of the previous task.

\begin{table}
    \centering
    \caption{Selection results on both test sets.}
    \label{table:selection}
	\begin{tabular}{l|rr|rr}
		\hline
										 & \multicolumn{2}{l|}{validation} & \multicolumn{2}{l}{test} \\ 
				 				  		 & R@1 & R@5 & R@1 & R@5 \\ \hline \hline
		baseline (ours)            		 & 91.9 & \textbf{99.7} & 91.0 & \textbf{99.3} \\ \cline{2-5}
		 - w/o long context			     & 81.4 & 97.9 & 84.0 & 97.7\\ \cline{2-5}
		 - w/o domain in input		         & 71.8 & 92.8 & 65.2 & 81.4 \\ \hline
		Multi-Task                		 & 94.3 & \textbf{99.7} & 81.5 & 97.9 \\ \hline
		Hierarch$_\text{domain+entity,doc}$& 96.3 & 98.1 & \textbf{93.2} & 97.3 \\ \cline{2-5}
		 - w/ multitask					 & 87.1 & 89.0 & 84.1 & 90.5 \\ \hline
		Hierarch$_\text{domain,entity,doc}$& \textbf{96.8} & 98.6 & 88.1 & 91.1 \\ \cline{2-5}
		 - w/ multitask					 & 95.7 & 97.7 & 86.2 & 88.5 \\ \hline
		DKR Triplet               		 & 90.9 & 98.9 & 87.2 & 96.9 \\ \cline{2-5}
		 - w/o RAG				      		 & 85.6 & 97.0 & 82.6 & 93.3 \\ \hline
		DKR Triplet hard          		 & 90.8 & 98.9 & 83.8 & 95.2 \\ \cline{2-5}
		 - w/o RAG					 		 & 88.0 & 97.6 & 84.7 & 95.8 \\ \hline
		DKR NLL                   		 & 93.4 & 98.8 & 85.5 & 96.8  \\ \cline{2-5}
		 - w/o RAG          		 		 & 90.1 & 98.1 & 84.0 & 94.4 \\ \hline
	\end{tabular}
\end{table}

\begin{table*}
    \centering
    \caption{Comparison of responses generated by RAG and the baseline method.}
    \label{table:rag}
    \begin{tabular}{p{0.15\linewidth}p{0.25\linewidth}p{0.48\linewidth}}
    \toprule[.1em]
    Last turn $u_T$ & selected snippets & Model: Generated response $u_{T+1}$\\
    \midrule[.1em]
    \multirow{2}{=}{Is their \colorbox{LightSpringGreen}{laundry} or \colorbox{LightSkyBlue}{dry cleaning} offered onsite?} & \small $k_1$: There are \colorbox{LightSpringGreen}{iron facilities} available, but no \colorbox{LightSpringGreen}{laundry facilities}. & \emph{baseline}: \small There are \colorbox{LightSpringGreen}{iron facilities} available, but no \colorbox{LightSpringGreen}{laundry facilities}. Is there anything else I can help you with?\\
      \addlinespace 
      & \small $k_2$: Allenbell does not provide \colorbox{LightSkyBlue}{dry cleaning} service. & \emph{RAG}: \small No, \colorbox{LightSkyBlue}{dry cleaning} is not offered onsite. There are \colorbox{LightSpringGreen}{iron facilities} available, but no \colorbox{LightSpringGreen}{laundry facilities}. \small Anything else I can do for you?\\    
	\midrule[.1em]
    \multirow{2}{=}{Do any of the \colorbox{LightSpringGreen}{trains} allow smoking?} & \small $k_1$: No, smoking is not allowed \colorbox{LightBrickRed}{at the stations}. & \emph{baseline}: \small No, smoking is not allowed \colorbox{LightBrickRed}{at the stations}. Anything else I can do for you? \\
    & \small $k_2$: No, smoking is not allowed \colorbox{LightSpringGreen}{on the train}. & \emph{RAG}: \small No, smoking is not allowed \colorbox{LightSpringGreen}{on the train}. Would you like to make a reservation? \\
    \bottomrule[.1em]
    \end{tabular}
\end{table*}

\begin{table}
    \caption{Comparison of different pre-trained transformer models for the baseline method of the selection task.}
    \label{table:lm_comparision}
    \centering
    \begin{tabular}{l|r|rrr}
        \hline
        &     & MRR &   & \\
        &  Params   & @5 & R@1   & R@5 \\ \hline \hline
        GPT-2  &       117M   & 91.9 & 87.2 & 97.6 \\ \hline
		GPT-2 medium & 345M   & 92.2 & 87.3 & 97.9 \\ \hline
		BART base	 &  139M     	& 87.0 & 77.1 & 99.2 \\ \hline
        BART large   &  406M      & \textbf{95.7} & 91.9 & 99.7 \\ \hline
		BERT large   & 335M   & 92.2 & 86.2 & 99.3  \\ \hline
        RoBERTa large& 355M	& \textbf{95.7} & \textbf{92.1} & \textbf{99.8} \\ \hline
    \end{tabular}
\end{table}

\subsubsection{Pre-Trained Model}
We evaluated a set of different pre-trained transformer models in different size variants on the baseline approach for the knowledge selection task.
The results can be found in \Cref{table:lm_comparision}.
In general, we observe better results by using the larger variants of the models.
Additionally, RoBERTa and BART seem to outperform GPT-2 and BERT on this task.
Since the differences between RoBERTa and BART are marginal and BART can also be used for the generation task in addition to the classification tasks, we decided to use BART large as our standard model.

\subsubsection{Domain}
As mentioned in \Cref{sec:method_baseline}, we include the domain of the knowledge snippet into its input representation to our models.
As shown in \Cref{table:selection} (w/o domain in input) this results in significant improvements in all three metrics for this task.
One of the main effects is that the confusion between domains with similar documents is reduced.
For example, the number of confusions between the taxi and train domain are reduced from 144 to 2.

\begin{table}
    \caption{Comparison of different decoding methods on the validation data.}
    \label{table:decoding_methods}
    \centering
    \begin{tabular}{l|rrr}
        \hline
        Method            & BLEU-1 & METEOR   & ROUGE-L \\ \hline \hline
        Sampling          & 41.6 & 42.1 & 40.9 \\ \hline
		Greedy Search     & 44.8 & 45.7 & \textbf{44.4} \\ \hline
		Beam Search	      & \textbf{45.3} & \textbf{46.8} & 44.2 \\ \hline
        - Beam Size 10    & 44.6 & 45.9 & 43.3 \\ \hline
    \end{tabular}
\end{table}

\subsubsection{Generation}
\Cref{table:decoding_methods} compares different decoding methods on the validation data.
We observe that beam search with a small beam size of four produces the best results compared to nucleus sampling, greedy search and a larger beam size of ten.

\subsubsection{Dialog Context Length}
As illustrated by \citep{kimDomainAPIsTaskoriented2020}, a long dialog context may often be beneficial when retrieving relevant knowledge. 
Consider for example a conversation, in which the entity in question in turn $u_T$ is only mentioned explicitly much earlier in the dialog context $u_1^{T-1}$.
Our experiments show that accounting for a longer context length significantly improves performance. 
As shown in \Cref{table:selection}, with the baseline model, we were able to increase performance on validation by 10.5\% absolute for R@1 by increasing the considered context length from 128 (w/o long context) to 384 tokens (baseline (ours)), which was the maximal feasible size on our GPUs.

\subsection{Negative Results}

\subsubsection{Multi-Task Training}
While some of our results using multi-task training appeared promising in the beginning, we did not see any major improvements overall, even when learning similar tasks jointly.
When training all three tasks of the challenge together, we only saw slight improvements in turn detection but degradation in both other tasks.
Similarly, when training our hierarchical models for knowledge selection we saw much better performance when all stages where trained on a single task, as illustrated in \Cref{table:selection}.
Furthermore, while our relevance classification model trained with auxiliary losses, where we extended our baseline selection model by auxiliary task-specific heads for domain, entity and document prediction, yielded decent improvements on the validation data, it failed to generalize to unseen data on test with strong performance degradation when compared to the single-task model.

\subsubsection{Retrieval Augmented Detection}
One general flaw with how we and \citet{kimDomainAPIsTaskoriented2020} approach the selection task is that the knowledge of what can be handled by the API and what not is learned from the training data.
Thus, in general, the system is not capable to adapt to new APIs of unknown domains or changes to known domains.
Since we assumed that every turn is either manageable by an API or by an unstructured knowledge document, an alternative way to approach the detection is to check whether a knowledge snippet exists which is relevant to the dialog.
This motivated us to apply the same approach as for the Retrieval Augmented Generation to the detection task.
Therefore, we marginalize our detection probability over \(K\) and apply the same approximations as in \Cref{sec:rag}.
The detection model thus receives the dialog context and one knowledge snippet as input.

We mainly tested this method on our artificial out of domain corpus but as shown on the results on the test set in \Cref{table:rac}, we did not see any significant improvements on top of the baseline method.
In contrast, we saw a significant degradation in the recall.
Initially, we assumed that this could be caused by cascaded selection errors, but a closer look at the results did not confirm this hypothesis.
Though theoretically, we think that the general approach is promising, we did not follow up further in this direction.

\subsubsection{Hard Batching}
Even though hard batching yielded improvements for the embedding models trained with triplet loss when compared to standard batching, the improvements were only minor.
Combining hard batching with fine-tuning the model with RAG even degraded the performance on the test set.
Generally, we saw better improvements using RAG, where the selection of the top \(n\) relevant snippets provides an implicit form of training on hard examples.
A possible explanation is that hard batching treats all but the ground truth knowledge snippet, of which there was always exactly one for each sample, as negatives. 
However, as seen in \Cref{table:rag}, multiple snippets might provide relevant knowledge.
The form of batching employed through retrieval augmentation appears to be more consistent, as multiple knowledge snippets can be deemed relevant and only parts, which would not contribute to a correct answer, would be penalized.

\begin{table}
    \caption{Results of Retrieval Augmented Detection (RAD). Both methods use BART base.}
    \label{table:rac}
    \centering
    \begin{tabular}{l|rrr}
        \hline
                    & Prec  & Rec   & F1 \\ \hline \hline
        baseline	& 99.5 & \textbf{92.8} & \textbf{96.0} \\ \hline
        Retrieval Augmented Detection	        & \textbf{99.6} & 91.5 & 95.5 \\ \hline
    \end{tabular}
\end{table}

\section{Conclusion}

We proposed two alternative methods for the knowledge selection task.
First, with Hierarchical Selection, we achieve our best results on the validation and test data and get a speedup of 24x compared to the baseline method.
Since the latency is still quite high with an average of 4.6 seconds, we propose Dense Knowledge Retrieval as another selection method.
It achieves an additional speedup, in total more than 2,500x compared to the baseline.
On the validation data, it even outperforms the baseline with respect to MRR@5 and R@1, but does not seem to generalize as well to the new modalities and domains in the test set.
Finally, we show that retrieval augmented generation can be used to consider multiple knowledge snippets and to jointly train selection and generation. 

As shown by other teams in this challenge there is still room for improvement with respect to performance. 
The detection task could benefit from tighter integration with the API-based dialog system, i.e. by joint modeling of dialog states and knowledge of the database schema.
Data augmentation could help to improve the generalizability of the different methods and to avoid overfitting.

\bibliography{literature.bib,literature_other.bib}

\end{document}